\documentclass[onecolumn,11pt]{article}
\usepackage[top=1in, bottom=1in, left=1in, right=1in]{geometry}
\usepackage[utf8]{inputenc}
\usepackage{amsmath,amsfonts,amscd,amssymb}
\usepackage{amsmath,bm}
\usepackage{bbm}
\usepackage{graphicx}
\usepackage{epstopdf}
\usepackage{overpic}
\usepackage{cancel}
\usepackage{rotating}
\usepackage{url}
\usepackage{color}
\usepackage{rotating}
\usepackage{multirow}
\usepackage{wrapfig}
\usepackage{mathtools}
\usepackage{subeqnarray}
\usepackage{setspace}
\usepackage{xcolor}
\usepackage{lscape}
\usepackage{rotating}
\usepackage[flushleft]{threeparttable}
\usepackage{caption}
\usepackage{booktabs}
\AtBeginEnvironment{sidewaystable}{}
\usepackage{subfiles}
\usepackage{makecell}
\usepackage{tabularx}
\usepackage{multicol}
\usepackage{hyperref}
\usepackage{etoolbox}
\usepackage{eso-pic}
\usepackage{palatino} 
\usepackage[numbers,sort&compress]{natbib}
\usepackage{algorithm}
\usepackage{algpseudocode}
\usepackage{titlesec}
\usepackage{chngcntr}
\usepackage{fancyhdr}
\usepackage{graphicx}

\usepackage[bottom,flushmargin,hang,multiple]{footmisc}
\usepackage{lipsum}
\newcommand\blfootnote[1]{%
  \begingroup
  \renewcommand\thefootnote{}\footnote{#1}%
  \addtocounter{footnote}{-1}%
  \endgroup
}

\usepackage{hyperref}
\hypersetup{
    colorlinks=true,
    linkcolor=blue,
    filecolor=magenta,      
    urlcolor=blue,
    citecolor=black,
    }

\definecolor{header1}{cmyk}{0,0,0,1}

\usepackage{listings}
\usepackage{xcolor}

\definecolor{codegreen}{rgb}{0.0,0.3,0.0}
\definecolor{codegray}{rgb}{0.5,0.5,0.5}
\definecolor{codepurple}{rgb}{0.1,0,0.82}
\definecolor{backcolour}{rgb}{1,1,1}

\lstdefinestyle{mystyle}{
    backgroundcolor=\color{backcolour},   
    commentstyle=\color{codegreen},
    keywordstyle=\bf\color{black},
    numberstyle=\tiny\color{codegray},
    stringstyle=\color{codepurple},
    basicstyle=\ttfamily\small,
    breakatwhitespace=true,         
    breaklines=true,                 
    captionpos=t,                    
    keepspaces=true,                 
    numbers=none,
    frame=lines,
    numbersep=5pt,                  
    showspaces=false,                
    showstringspaces=false,
    showtabs=false,                  
    tabsize=3
}

\usepackage{mathabx}
\usepackage{subcaption}

\usepackage[utf8]{inputenc}

\title{\vspace{-.55in}{\fontsize{17.5}{17.5}\selectfont\textbf{Sparse Identification of Nonlinear Dynamics with Conformal Prediction}}

\author{\normalsize{Urban Fasel}\\
\footnotesize{Department of Aeronautics, Imperial College London, UK}\vspace{-.1in}}
}

\date{}
\begin{document}
\maketitle

\lstdefinestyle{interfaces}{
  float,
  floatplacement=tbp
}

\blfootnote{Code \url{https://github.com/urban-fasel/SINDyCP_tutorials}.}
\vspace{-.2in}
\begin{abstract}

The Sparse Identification of Nonlinear Dynamics (SINDy) is a method for discovering nonlinear dynamical system models from data. Quantifying uncertainty in SINDy models is essential for assessing their reliability, particularly in safety-critical applications. 
While various uncertainty quantification methods exist for SINDy, including Bayesian and ensemble approaches, this work explores the integration of Conformal Prediction, a framework that can provide valid prediction intervals with coverage guarantees based on minimal assumptions like data exchangeability. 
We introduce three applications of conformal prediction with Ensemble-SINDy (E-SINDy): (1) quantifying uncertainty in time series prediction, (2) model selection based on library feature importance, and (3) quantifying the uncertainty of identified model coefficients using feature conformal prediction. We demonstrate the three applications on stochastic predator-prey dynamics and several chaotic dynamical systems. We show that conformal prediction methods integrated with E-SINDy can reliably achieve desired target coverage for time series forecasting, effectively quantify feature importance, and produce more robust uncertainty intervals for model coefficients, even under non-Gaussian noise, compared to standard E-SINDy coefficient estimates.

\end{abstract}

\section{Introduction}

The Sparse Identification of Nonlinear Dynamics (SINDy) is a method for discovering governing equations of dynamical systems purely from data~\cite{brunton2016discovering}. The method has been successfully applied across a range of scientific and engineering domains, including fluid dynamics, neuroscience, and epidemiology~\cite{fukami2021sparse, delabays2025hypergraph, horrocks2020algorithmic}.
Quantifying uncertainty in SINDy models is essential for assessing the reliability of the discovered dynamics, especially in safety-critical applications like autonomous driving, robotics, or medical decision-making~\cite{drgona2025safe}. Uncertainty quantification (UQ) in the context of SINDy can include estimating uncertainty in the model coefficients, accounting for uncertainty in the structure of the learned equations, and propagating uncertainty during forecasting through numerical integration of the learned dynamical system. 
A variety of methods have been proposed for UQ in SINDy: Bayesian approaches offer posterior distributions over model coefficients~\cite{zhang2018robust,yang2020bayesian,galioto2020bayesian,hirsh2022sparsifying,north2022bayesian,niven2024dynamical,fung2025rapid,klishin2024statistical}, ensemble-based methods average over multiple models to account for data and model variability~\cite{fasel2022ensemble,maddu2022stability,gao2023convergence}, while filtering and data assimilation approaches allow for online updating of model estimates and uncertainties~\cite{rosafalco2024ekf,rosafalco2025online}. Recent work has also introduced a variational formulation of SINDy to jointly learn dynamics and uncertainty within a reduced-order probabilistic framework~\cite{conti2024veni}.
Conformal prediction, a relatively new framework for uncertainty quantification, provides model-agnostic prediction intervals with coverage guarantees, offering a promising addition to existing SINDy-UQ techniques by ensuring rigorous confidence levels regardless of the underlying model assumptions~\cite{vovk2005algorithmic}.

Conformal prediction is a framework for quantifying the uncertainty of predictions made by any algorithm, providing prediction intervals that are statistically guaranteed to contain the true value with a specified confidence level~\cite{vovk2005algorithmic}. This is achieved by evaluating how well new data conforms to previously observed data, and using this information to construct prediction intervals that reflect the uncertainty in the model’s predictions. The method works under the assumption that the data is exchangeable, meaning the joint distribution of the data points or random variables is invariant under permutations. This is a weaker assumption than the data being independent and identically distributed (i.i.d.), only assuming that the order of the data points does not affect their statistical properties~\cite{vovk2005algorithmic,angelopoulos2024theoretical}. However, exchangeability is often violated in time series data, where observations are temporally dependent and the order of data points is crucial.
To address this challenge, various extensions of conformal prediction have been developed. For instance, Ensemble-based Prediction Intervals (EnbPI) adapts conformal prediction for time series forecasting~\cite{xu2021conformal}, enabling uncertainty quantification in dynamical systems predictions. Similarly, the Conformal PID Control method combines principles from control theory and conformal prediction to provide reliable uncertainty quantification in time series forecasting~\cite{angelopoulos2023conformal}. 
Conformal prediction has also been extended for covariate importance measure in model selection, with approaches like LOCO (Leave-One-Covariate-Out) \cite{lei2018distribution} and LOCO-path \cite{cao2023inference}. These methods offer new ways to calculate variable importance that can be applied in sparse model discovery methods like SINDy. 
Another interesting conformal prediction extension is feature conformal prediction that quantifies the uncertainty associated with individual features, deploying conformal prediction in the feature space rather than the output space~\cite{teng2022predictive,chen2024conformalized}. Applied to SINDy, this can help quantify uncertainty in the identified model coefficients and can help assess which candidate nonlinear terms are reliably contributing to the dynamics, potentially improving robustness in sparse model selection.

\begin{figure}[t]
    \centering
    \includegraphics[width=1.0\textwidth]{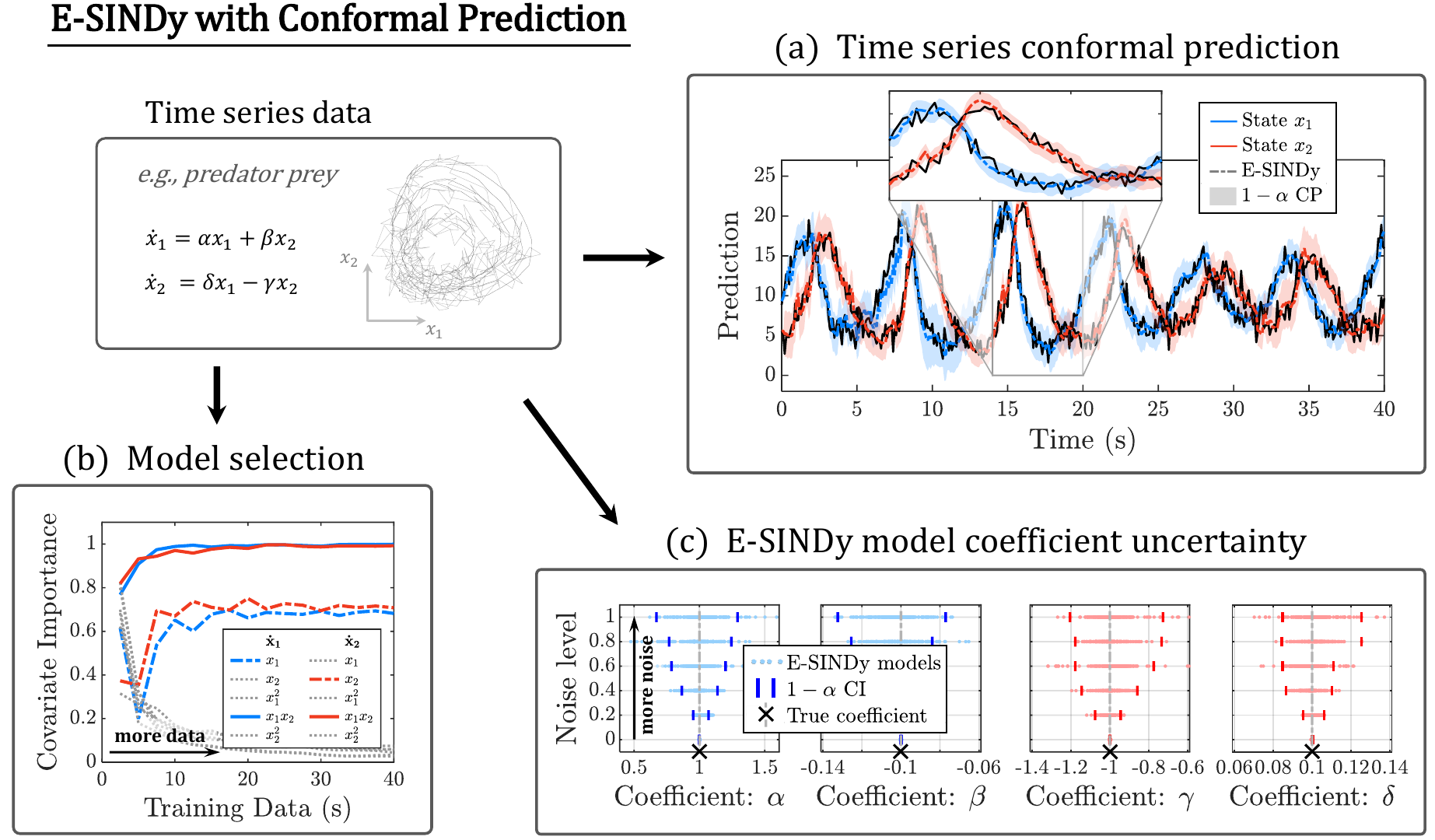}
    \vspace{2pt}
    \caption[ ] {Schematic of the E-SINDy with conformal prediction framework for (a) uncertainty quantification in time series prediction, (b) model selection using library feature (covariate) importance measures, and (c) quantifying uncertainty in the identified model coefficients. \vspace{6pt}}
    \label{fig1}
\end{figure}

In this paper, we explore methods that combine conformal prediction with the SINDy framework. An overview of the methods is shown in Figure~\ref{fig1}. While not intended as a comprehensive review, we aim to highlight promising strategies for quantifying uncertainty in SINDy’s model predictions, estimated coefficients, and identified model structures. We begin with a short overview of SINDy and its ensemble variant, and then introduce the conformal prediction techniques that form the core of this paper. To demonstrate these methods in practice, we apply them to a simple two-dimensional stochastic predator-prey system and several more challenging chaotic dynamical systems. All code and examples are available on \url{https://github.com/urban-fasel/SINDyCP_tutorials}.

\section{Background} \label{sec:bg}

Here, we introduce the SINDy and conformal prediction methods. We first introduce the SINDy algorithm~\cite{brunton2016discovering}, followed by its ensemble extension~\cite{fasel2022ensemble}. Second, we introduce conformal prediction, in particular split conformal prediction and its variants~\cite{lei2018distribution}, followed by conformal prediction extensions for time series data~\cite{xu2021conformal,angelopoulos2023conformal}.

\subsection{Sparse Identification of Nonlinear Dynamics (SINDy)} \label{sec:SINDy}

The SINDy algorithm is a data-driven approach to discovering governing equations of dynamical systems directly from time series data~\cite{brunton2016discovering}. Given an observation of the system state $\mathbf{x}(t) \in \mathbb{R}^{m}$ evolving in time $t$, the goal is to discover a dynamical system model $\mathbf{f}$ that best fits the data:
\begin{equation}
    \frac{d}{dt}\mathbf{x}(t) = \mathbf{f}(\mathbf{x}(t)).
\end{equation}
The algorithm uses sparse regression to identify a small set of active terms from a predefined library of candidate functions that best describe the observed time series data. 
Given a time series of $n$ state snapshots, we can stack the snapshots to form a time series data matrix $\mathbf{X} \in \mathbb{R}^{n \times m}$, where each row is a time snapshot and each column represents a state variable. With measured or numerically calculated time derivatives $\frac{d}{dt}\mathbf{x}(t) = \dot{\mathbf{x}}(t) \in \mathbb{R}^{m}$, we form a data matrix of time derivatives $\dot{\mathbf{X}} \in \mathbb{R}^{n \times m}$. SINDy then finds a sparse model that describes the dynamics of the form:
\begin{equation}
    \dot{\mathbf{X}} = \boldsymbol{\Theta}(\mathbf{X}) \mathbf{\Xi},
\end{equation}
where $\boldsymbol{\Theta}(\mathbf{X}) \in \mathbb{R}^{n \times p}$ is a library of $p$ candidate nonlinear functions (e.g., polynomials) evaluated at the $n$ observed state data snapshots, and $\mathbf{\Xi} \in \mathbb{R}^{p \times m}$ is a matrix of unknown coefficients, with each column representing the terms that contribute to the dynamics of the corresponding state variable. 
The key idea of SINDy is that for many systems, each state’s dynamics depend only on a small subset of terms in the library. Thus, the estimation of $\mathbf{\Xi}$ is formulated as a sparse regression problem: 
\begin{equation}
    \mathbf{\Xi} = \mathrm{argmin}_{\hat{\boldsymbol{\Xi}}} \|\dot{\mathbf{X}} - \boldsymbol{\Theta}(\mathbf{X})\hat{\boldsymbol{\Xi}}\|_2^2 + \lambda\|\hat{\boldsymbol{\Xi}}\|_0.
\end{equation}
One common strategy to approximately solve the $L_0$ regularized least squares problem is sequential thresholded least squares (STLSQ), which alternates between two steps. First, the least-squares problem: $\mathbf{\Xi} = \mathrm{argmin}_{\hat{\boldsymbol{\Xi}}} \|\dot{\mathbf{X}} - \boldsymbol{\Theta}(\mathbf{X})\hat{\boldsymbol{\Xi}}\|_2^2$ is solved. Then, small coefficients in $\mathbf{\Xi}$ (below a chosen threshold $\lambda$) are set to zero, and the least-squares problem is solved again using only the remaining active terms.
The threshold parameter $\lambda$ controls the sparsity of the final model and can be selected via cross-validation, information criteria, stability selection, or domain knowledge.  

\paragraph{Ensemble SINDy} SINDy identifies dynamical system models that are sparse, interpretable, and robust to moderate noise in the observations. However, under high noise or limited data, the performance of SINDy can degrade, motivating extensions such as the ensemble~\cite{fasel2022ensemble} or weak-form~\cite{messenger2021weak,reinbold2020using} approaches. 
Ensemble-SINDy (E-SINDy) is an extension of the original SINDy method, designed to improve the robustness of model discovery in the presence of noise and limited data~\cite{fasel2022ensemble}. The key idea behind E-SINDy is to incorporate ensemble learning into the SINDy framework, using multiple models derived from different bootstrapped samples of the data. This ensemble approach helps to improve the noise robustness and introduces a way to quantify the uncertainty of the identified model.  
The method follows four general steps. (1) Data resampling: first, $B$ bootstrap samples from the original dataset are generated. These samples are created by randomly sampling rows with replacement from the time derivatives and library matrix $[\dot{\mathbf{X}}, \boldsymbol{\Theta}(\mathbf{X})]\in \mathbb{R}^{m+p \times n}$, leading to multiple resampled datasets that capture potential variations in the underlying system dynamics. (2) Model fitting: for each of the $B$ bootstrap samples, the standard SINDy procedure to discover a model of the system dynamics is applied. This results in $B$ different SINDy models, each with its own set of coefficients and candidate terms. (3) Coefficient aggregation: after identifying the individual models, the coefficients across all ensemble members are aggregated. Common strategies for aggregation include computing the median or mean of the coefficients. This step helps to identify terms that consistently appear across the ensemble of models, increasing confidence in their relevance to the underlying dynamics. (4) Thresholding: last, a sparsity threshold can be applied to the aggregated coefficients to eliminate terms that have a low ensemble inclusion probability and are considered irrelevant across the ensemble. 

By examining the variability of the coefficients and the consistency of the selected terms across the ensemble, E-SINDy provides a mechanism for quantifying uncertainty in the discovered models. This enables more informed predictions and enhances model confidence, thereby improving the standard SINDy method by providing more reliable model discovery. 

\subsection{Conformal Prediction}

Conformal prediction is a framework for quantifying uncertainty in machine learning models by generating valid prediction intervals. The only assumption regarding data distribution is exchangeability. Conformal prediction works by assigning a non-conformity score to each data point, quantifying how different a point is from the rest of the dataset. Conformal prediction ensures that the predicted intervals contain the true outcome with a specified probability, even if the underlying data distribution is unknown.

For a regression problem with i.i.d. data $ (X_1, Y_1), \dots, (X_n, Y_n) \sim P$, each random variable $ (X_i, Y_i) \in \mathbb{R}^p \times \mathbb{R} $ comprises of features (also called predictors or covariates) $X_i = (X_i(1) \dots, X_i(p))$ with feature dimension $p$ and responses $Y_i$. We denote $(X_i, Y_i)$ random variables and $(x_i,y_i)$ data samples that are realizations of $(X_i, Y_i)$. In the SINDy context, the features $X_i$ would correspond to the library features $\boldsymbol{\Theta}(\mathbf{x}_i) \in \mathbb{R}^p$ at time step $i$ (with $p$ candidate nonlinear functions), and the response $Y_i$ would be the time derivatives $\dot{\mathbf{x}}_i$. For an unknown regression function $f(x), x \in \mathbb{R}^p$, the goal of conformal prediction is to predict a new response $ Y_{n+1} $ for a new input $ X_{n+1} $, and build a prediction interval $ C_\alpha(X_{n+1}) $ that contains the true value $ Y_{n+1} $ with high probability and small interval width. More specifically, for a user-specified error rate $ \alpha \in [0,1] $, we want:
\begin{equation}
    \mathbb{P}(Y_{n+1} \in C_\alpha(X_{n+1})) \ge 1 - \alpha.
\end{equation}
This requirement, referred to as \textit{marginal coverage}, ensures that the actual response $ Y_{n+1} $ falls within the predicted interval $ C_\alpha(X_{n+1}) $ with probability at least $ 1 - \alpha $, and we want to guarantee this without any assumptions on the regression function $f$ and data distribution $P$, except for the data points to be exchangeable.

\paragraph{Split Conformal Prediction} A range of methods exists to build conformal prediction intervals that guarantee coverage, with different tradeoffs
between computational efficiency and width of resulting prediction intervals. Split Conformal Prediction (SCP)~\cite{lei2018distribution} is a computationally efficient method that builds the interval $ C_\alpha(X_{n+1}) $ by first splitting the data into two equal-sized sets, a training set $\mathcal Tr$ and a calibration set $\mathcal Cal$. On the training set, a regression model $ \hat f(x) $ is fit (which could be a SINDy model), and then used on the calibration set to build the prediction intervals using a non-conformity score $s(x,y)$. For regression problems, the absolute value of the residuals is often used as a non-conformity score:
\begin{equation}
    s(x_i, y_i) = |y_i - \hat f(x_i)|.
\end{equation}
Those scores are taken to define the prediction interval using the empirical quantile $q_{1 - \alpha}$ defined as $\text{the } k\text{th smallest value in } \{s_i : i \in \mathcal Cal\}\text{, where } k = \lceil (n/2 + 1)(1 - \alpha) \rceil$: 
\begin{equation}
    C_\alpha(x_{n+1}) = [\hat{f}(x_{n+1}) - q_{1 - \alpha}, \hat{f}(x_{n+1}) + q_{1 - \alpha}].
\end{equation}
SCP is a computationally efficient conformal prediction method, but splitting the training and calibration sets introduces some randomness to the process. A common way to reduce this randomness is to combine the results from multiple different splits of the data, or to use jackknife prediction that uses the quantiles of leave-one-out residuals to define the prediction intervals~\cite{lei2018distribution}. 
The jackknife method improves on SCP by using more of the data when computing the non-conformity score and quantiles, often resulting in shorter prediction intervals. However, its intervals do not guarantee coverage in finite samples. The jackknife+ extension addresses this and achieves valid coverage for exchangeable data regardless of the underlying data distribution~\cite{barber2021predictive}, and can also be applied to conformal prediction with ensemble methods~\cite{kim2020predictive}.

\paragraph{Time Series Data} A fundamental assumption of conformal prediction is that the data is exchangeable. This assumption is often violated in time series data, where temporal dependencies exist. Many recent papers discuss extensions of conformal prediction to work in this setting~\cite{xu2021conformal,gibbs2021adaptive,zaffran2022adaptive,angelopoulos2023conformal,auer2023conformal}. Two interesting methods are the Ensemble batch prediction interval method~\cite{xu2021conformal} and the Conformal PID Control~\cite{angelopoulos2023conformal} method.
The \textbf{Ensemble batch Prediction Interval (EnbPI)} method constructs prediction intervals by using the residuals from ensemble models. For each time point, it computes the residuals using only those models that did not include the time point in their training data, ensuring out-of-sample error estimation. These residuals are aggregated to form a distribution, from which quantiles are selected to define the prediction intervals, thereby providing coverage guarantees without relying on strong distributional assumptions.
The second method, \textbf{Conformal PID Control (CP-PID)}, combines principles from control theory and conformal prediction to provide reliable prediction intervals in time series forecasting. The method adjusts prediction intervals based on feedback from past prediction errors, similar to how a PID (Proportional-Integral-Derivative) controller adjusts control inputs based on error signals. The prediction interval at each time step is again determined by a quantile of past non-conformity scores (e.g., absolute prediction errors). The PID controller updates this quantile dynamically to maintain the desired $(1-\alpha)$ coverage level. This adaptive mechanism allows the method to respond to changes in the data distribution, such as trends or seasonality, ensuring that the prediction intervals remain valid over time. 

For both methods, guaranteeing marginal coverage without making any assumptions about the data sequence is challenging. The prediction intervals by EnbPI provide approximately valid marginal coverage under mild assumptions on the regression estimators and the stochastic errors of the time series. For the CP-PID method, long-run coverage over time is achieved only assuming boundedness of the non-conformity scores. For large values of time $T$, and under minimal assumptions, the average miscoverage rate is expected to approach the target level $\alpha$:
\begin{equation}
    \frac{1}{T} \sum_{t=1}^{T} \text{err}_t 
 = \frac{1}{T} \sum_{t=1}^{T} \mathbbm{1} \{ y_t \notin C_t \} = \alpha + o(1),
\end{equation}
where $o(1)$ represents a term that converges to zero as $T \rightarrow \infty$.

\vspace{14pt}

\section{Methods} \label{sec:method}

Here, we introduce E-SINDy with conformal prediction. We incorporate conformal prediction with E-SINDy in three different ways: (1) for uncertainty quantification in time series prediction, (2) model selection using library feature importance measures, and (3) quantifying the uncertainty of identified model coefficients using surrogate feature conformal prediction. Full algorithmic details and code to reproduce the result can be found in the accompanying \href{https://github.com/urban-fasel/SINDyCP_tutorials}{GitHub tutorial}.

\subsection{Time series prediction}

\paragraph{EnbPI with E-SINDy}

We first introduce the EnbPI method with E-SINDy for constructing prediction intervals in time series settings. Given time series data, EnbPI with E-SINDy can produce prediction intervals that adapt to distribution shifts and non-stationarity in the data. The method is useful for real-time applications due to its computational efficiency and theoretical coverage guarantees.
We begin by collecting time series training data, recording the system state $\{\mathbf{x}_t\}_{t=1}^{n}$ over $n$ time steps. From this data, we construct a library of candidate features and estimate time derivatives via finite difference or a weak form approach~\cite{messenger2021weak,reinbold2020using}.
Next, we train an ensemble of $B$ SINDy time series forecasting models, $\hat{f}_1, \ldots, \hat{f}_B$, using bootstrap resampling of the training set. Each forecasting model $\hat{f}_b$ includes the identified SINDy differential equation, a numerical integrator, and a state estimator. The numerical integration method we use is Runge-Kutta 4th order, and as a state estimator, we simply smooth the noisy measurements using a Savitzky–Golay filter~\cite{savitzky1964smoothing}, although more sophisticated smoothing~\cite{rudy2019smoothing} or data assimilation methods~\cite{rosafalco2024ekf,asch2016data} might improve the performance of the method. The models $\hat{f}_b$ then take a measurement of the current system state $\mathbf{x}_n$ (or a sequence of past measurements) as input, and predict the future state trajectory $\{\mathbf{\hat x}_t\}_{t=n+1}^{n+t_p}$ over a prediction horizon of $t_p$ time steps.
To calculate the non-conformity scores $s_t$, we compute residuals within a sliding calibration window of size $l_r$. For each time $t$ in this window, we evaluate residuals using only those models that were not trained on the corresponding state $\mathbf{x}_t$, ensuring that errors are estimated out-of-sample. Specifically, for each model $b$ in the ensemble:
\begin{equation}
    s_t^{(b)} = |y_t - \hat{f}_b(\mathbf{x}_t)|, \quad b = 1, \dots, B,
\end{equation}
where $y_t = {\{\mathbf{x}_i\}_{i=t+1}^{t+t_p}}$. We then aggregate these residuals across the ensemble to obtain a pointwise residual $s_t = \frac{1}{B} \sum_{b=1}^B s_t^{(b)}(\mathbf{x}_t)$. 
With this collection of non-conformity scores $\mathcal{S} = \{s_t\}_{t=n-l_r}^{n-1}$, we estimate the $(1 - \alpha)$ quantile:
\begin{equation}
    q_{1-\alpha} = \text{Quantile}_{1-\alpha}(\mathcal{S}).
\end{equation}
This quantile allows us to construct prediction intervals for new observations. For a test input $\mathbf x_{n+1}$, the ensemble prediction $\bar{f}(\mathbf{x}_{n+1}) = \frac{1}{B} \sum_{b=1}^B \hat{f}_b(\mathbf{x}_{n+1})$ is computed, and the prediction interval is defined as:
\begin{equation}
    C_{n+1} = \left[\bar{f}(\mathbf{x}_{n+1}) - q_{1-\alpha},\; \bar{f}(\mathbf{x}_{n+1}) + q_{1-\alpha}\right].
    \vspace{4pt}
\end{equation}
By continuously recalibrating prediction intervals with newly collected measurement data, we allow the interval widths to adapt dynamically to trends or shifts in the underlying dynamics. When new data becomes available after the ensemble has been trained, it can be used with all existing models to update the residual quantile estimates. Moreover, due to the computational efficiency of training SINDy models, it is also feasible to update the E-SINDy model itself periodically, further adapting the forecaster to changing dynamics.

This approach offers several practical advantages that make it well-suited for real-time settings. First, it avoids the need to split the data into separate training and calibration sets, thereby allowing the entire dataset to be used both for model fitting and for uncertainty quantification. This maximizes data efficiency, which is especially beneficial when working with limited data. Second, the method is computationally efficient. The ensemble of SINDy models does not require retraining at the same frequency as the interval construction. Third, the method is adaptable to non-stationary or dynamically evolving environments. As new measurements are collected, the prediction intervals can be recalibrated in real time to reflect trends, seasonality, or changes in system behavior. Finally, under mild assumptions, the method guarantees coverage asymptotically, even when the data points are not exchangeable. This makes the method broadly applicable across a wide range of dynamical systems and forecasting tasks.

\paragraph{Conformal PID Control with E-SINDy}

The second method for uncertainty quantification in time series prediction uses E-SINDy with the conformal PID control method~\cite{angelopoulos2023conformal}. 
The Conformal PID Control framework combines principles from control theory and conformal prediction to provide reliable uncertainty quantification in time series forecasting. The method adjusts prediction intervals based on feedback from past prediction errors, similar to how a PID controller adjusts control inputs based on error signals.
To quantify uncertainty in time series prediction, the conformal prediction interval at each time step is determined by a quantile of past non-conformity scores. The PID controller updates this quantile dynamically to maintain the desired $(1-\alpha)$ coverage level. 
In control terms, the control variable is the quantile $q_t$ and the process variable is the miscoverage rate $ \mathrm{err}_t = \mathbbm{1} \{ y_t \notin C_t \}$. The objective is to control the process variable $\mathrm{err}_t$ around a desired set point $\alpha$. To do this, corrections are applied to the quantile $q_t$ based on the error of the output $g_t=\mathrm{err}_t-\alpha$ between the process variable and the target level.
Same as for EnbPI, this adaptive mechanism responds to changes in the data distribution, such as trends or seasonality, ensuring that the prediction intervals remain valid over time. 

The conformal PID control method with SINDy is initialized similarly to EnbPI with SINDy. First, we train an ensemble of SINDy models, used as base forecasters. In the conformal PID control method, the quantiles are then built using proportional (P), integral (I), and derivative (D) control. 
The P control tracks the quantile on the past quantile loss to maintain long-run coverage, assuming only that the score magnitudes are bounded. I control (or error integration) stabilizes coverage by adjusting predictions using the accumulated coverage error or running sum $\sum g_i=\sum_{i=1}^{t} (\mathrm{err}_i - \alpha)$ of the coverage errors. Finally, the D control (or scorecaster) introduces a model to predict the next score’s quantile, helping to adapt to trends or shifts in the data that would not be captured by the base forecaster. We do not use the scorecaster in our work, since we can update the base SINDy forecaster using incoming data. 
The P and I components used here define the conformal PI controller:
\begin{equation}
    q_{t+1} = \underbrace{\eta g_t}_{\text{P}} + \underbrace{r_t\left( \sum_{i=1}^t g_i \right)}_{\text{I}},
    \label{eq:pid}
\end{equation}
where $\eta$ is the P control gain, and $r_t$ is a saturation function~\cite{angelopoulos2023conformal}.

Similar to the EnbPI method, conformal PID control offers several practical advantages. It ensures long-run marginal coverage under minimal assumptions, without requiring exchangeability or stationarity of the data. It is also adaptive to shifts in the data through its feedback-based updating mechanism, which makes it suitable for changing environments, and the approach is computationally efficient, same as with EnbPI as it does not require model retraining at high frequency.

\subsection{Model selection: library feature importance}\label{modelSelection}

At the core of the SINDy framework is a model selection procedure using sparse regression to identify the most parsimonious model consistent with the observed data. 
In recent years, conformal prediction-inspired methods have been explored for quantifying uncertainty in model selection, such as in the Leave-One-Covariate-Out (LOCO) method~\cite{lei2018distribution}. 
The LOCO method is closely related to permutation feature importance~\cite{breiman2001random} and has conceptual similarities with stability selection~\cite{maddu2022stability} and thresholding methods based on ensemble inclusion probabilities~\cite {fasel2022ensemble}. 
In the following, we introduce two methods incorporating LOCO with SINDy.

\paragraph{Leave-One-Covariate-Out (LOCO)}

We apply the LOCO method in the context of the SINDy library to estimate the importance of each library feature. First, we identify a SINDy model ${\boldsymbol{\Xi}}$ trained on the full dataset $\{\mathbf{x}_i\}_{i=1}^n$ using all available features in the library. To evaluate the importance of the $j$th feature, we train the SINDy model after removing that feature, identifying a new model ${\boldsymbol{\Xi}}^{(-j)}$:
\begin{equation}
    \boldsymbol{\Xi}^{(-j)} = \mathrm{argmin}_{\hat{\boldsymbol{\Xi}}^{(-j)}} \|\dot{\mathbf{X}} - \boldsymbol{\Theta}(\mathbf{X})\hat{\boldsymbol{\Xi}}^{(-j)}\|_2^2 + \lambda\|\hat{\boldsymbol{\Xi}}^{(-j)}\|_0.
    \label{LOCO-SINDy}
    \vspace{4pt}
\end{equation}
For a new data point $\mathbf{x}_{n+1}$, we define the excess prediction error $\Delta_j(\mathbf{x}_{n+1})$ of the $j$th feature as: 
\begin{equation}
    \Delta_j(\mathbf{x}_{n+1}) = \|\dot{\mathbf{x}}_{n+1} - \boldsymbol{\Theta}(\mathbf x_{n+1}){\boldsymbol{\Xi}}^{(-j)}\|_1 - \|\dot{\mathbf{x}}_{n+1} - \boldsymbol{\Theta}(\mathbf x_{n+1}){\boldsymbol{\Xi}}\|_1.
\end{equation}
This value captures the increase in prediction error caused by excluding the $j$th feature and serves as a basis for inference on its predictive relevance.

To estimate this error, we use the jackknife+ resampling strategy. We construct an ensemble of SINDy models, each trained on all but one of the training points, left out for calculating the excess error $\Delta_j(\mathbf{x}_{n+1})$. Within each jackknife iteration, we refit a SINDy model excluding the $j$th feature to compute the excess error. Since the error magnitudes are not directly interpretable, we normalize them over all features to define an importance score for each feature. A higher score indicates greater importance compared to other features, as it shows a larger degradation in predictive accuracy when the feature is removed.

One limitation of the LOCO approach is its sensitivity to correlated features, underestimating the importance of correlated features, since other features can compensate for the missing information. This is discussed broadly in the permutation importance literature, where the problem is addressed by permuting covariates in subgroups~\cite{strobl2008conditional}. This can also be applied to the LOCO with SINDy approach, calculating LOCO importance measures on subgroups of library features, or grouping correlated features before analysis.
Another potential challenge with the method, which would even be exacerbated when subgroups are considered, is that it is computationally expensive, as it requires retraining the SINDy models within each jackknife resample and for each feature left out. Moreover, comparing models trained with different feature sets can introduce inconsistencies, making interpretation difficult.

\paragraph{LOCO-path} 

One additional challenge in the standard LOCO approach is that the regularization parameter $\lambda$ needs to be selected, as in most sparse regression methods. Typically, $\lambda$ is chosen through cross-validation, as is common in LASSO~\cite{tibshirani1996regression}, using information criteria like AIC~\cite{mangan2017model}, or via stability selection~\cite{maddu2022stability}. A recent extension of LOCO, originally proposed for the LASSO and adapted here to SINDy, avoids the need for $\lambda$ selection by evaluating variable importance across the entire regularization path~\cite{cao2023inference}. This approach, referred to as the LOCO solution path or short LOCO-path, involves computing the LOCO importance measure over a range of $\lambda$ values, thereby capturing the influence of a feature throughout the model’s regularization path.

We first define the LOCO solution path, which uses the set of identified SINDy models over the $\lambda$ path after removing feature $j$ from the library to calculate an importance measure. Compared to the SINDy LOCO importance measure, where we calculate an excess prediction error due to removing the covariate, here we use an estimated model coefficient error. The LOCO SINDy coefficient for the sparse regression with excluded library feature $j$ is obtained using Equation~\ref{LOCO-SINDy}. To quantify the difference introduced by excluding feature $j$, we define a statistic $T_j$, 
\begin{equation}    T_{j}=\sum_{\lambda_1}^{\lambda_k}\|\boldsymbol{\Xi}^{(-j)}(\lambda)-\boldsymbol{\Xi}(\lambda)\|_1.
\label{LOCO-path}
\end{equation}
which measures the cumulative deviation between the full model's solution ${\boldsymbol{\Xi}}(\lambda)$ and the LOCO SINDy solution ${\boldsymbol{\Xi}}^{(-j)}(\lambda)$ across a range of $\lambda$ values. This statistic serves as a regularization-path-aware importance score, reflecting how consistently important a variable is across different sparsity levels.

As with standard LOCO, a known challenge arises when features are correlated, as the effect of removing one feature can be compensated for by others. This limitation persists here, but can be addressed using similar extensions, such as grouping correlated features before performing the analysis.

\subsection{Model coefficient uncertainty estimation: feature-CP}

A final idea we explore here involves applying conformal prediction directly to the estimated model coefficients $\boldsymbol{\Xi}$ of the library features, instead of the SINDy response or future states prediction. 
The goal is to quantify uncertainty in the structure and values of the learned dynamical system. The idea is similar to Bayesian or ensemble learning for estimating model coefficient uncertainty, which is broadly discussed in the literature with a recent work on comparing uncertainty estimates in Bayesian and ensemble learning applied to sparse model discovery~\cite{gao2023convergence}. It was demonstrated that ensemble SINDy, especially when using bootstrapping-based least-squares combined with sequential thresholding, offers computationally efficient and statistically valid uncertainty quantification. As sample sizes increase, E-SINDy model coefficient estimates tend to concentrate around the true model coefficients. 
However, the confidence intervals around ensemble-based coefficient estimates are often narrow, especially with limited data, and may fail to contain the ground-truth values, e.g., evaluated on synthetic data benchmarks~\cite{gao2023convergence}. 
This can result from bias introduced by the thresholding step in SINDy, or from the dependence between bootstrap samples drawn with replacement, which can underestimate variability, especially in small samples or dependent data, such as in time series. 
Another issue is that SINDy’s least squares step assumes noise only in the output (time derivatives), ignoring noise in the inputs (library functions), although measurement noise affects both. Addressing this would require solving an errors-in-variables problem, as recently proposed in~\cite{bortz2023direct}. 

In this work, we explore conformal prediction to better quantify uncertainty in the identified model coefficients. A key challenge in this setting is the lack of a calibration dataset for the learned model coefficients, unless synthetic re-learning experiments are performed where a ground truth model would be available. 
To overcome this, surrogate features can be constructed to calibrate prediction intervals, an idea recently introduced as feature conformal prediction (feature-CP)~\cite{teng2022predictive}. 
Applied with SINDy, feature-CP first trains an ensemble of SINDy models using jackknife sampling, where each model is trained on all but one data point (or one batch of data points for larger datasets). For each excluded data point $\mathbf{x}_i$ (or batch of data points), a surrogate SINDy model is identified. 
For the case of using a single data point, the surrogate SINDy model can be identified by solving a constrained (sequentially thresholded) least-squares problem~\cite{loiseau2018constrained}: 
\begin{alignat}{2}
    & \min_{\boldsymbol{\xi}_k} && \quad \| {\boldsymbol{\Theta}}(\mathbf{X}^{(-i)}) {\boldsymbol{\xi}_k} - \dot{\mathbf{X}}^{(-i)}_k \|_2^2, \\
    & \ \text{subject to } && \quad {{\boldsymbol{\Theta}}(\mathbf{x}_i)}{\boldsymbol{\xi}_k} = {\dot{\mathbf{x}}_i},
\end{alignat} 
where $\boldsymbol \xi_k \in \mathbb{R}^{p}$ is the model coefficient vector of state $k=1, \dots, n$ of the sparse matrix of model coefficients $\boldsymbol \Xi$, with $p$ library features and $n$ states, e.g., $\boldsymbol{\Xi} = [\boldsymbol{\xi}_1, \boldsymbol{\xi}_2]$ for $n=2$.
The least-squares problem is solved with an additional linear equality constraint on the coefficients $\boldsymbol{\Xi}$ to reproduce the response $\dot{\mathbf{x}}_i$ exactly when evaluated on the excluded data. When using batches of data points, an additional SINDy problem would be solved on the excluded batch of data points, e.g., using the recent multi-objective SINDy~\cite{lemus2025multi}. The constrained least squares problem can be reformulated as an unconstrained optimization where the original constraints are enforced through the use of Lagrange multipliers. The optimal solution that satisfies the equality constraints is also the solution to the Karush-Kuhn-Tucker equations~\cite{loiseau2018constrained}:
\begin{equation}
    \begin{bmatrix}
    \boldsymbol{\Theta}(\mathbf{X}^{(-i)})^T \boldsymbol{\Theta}(\mathbf{X}^{(-i)}) & \boldsymbol{\Theta}(\mathbf{x}_i)^T \\
    \boldsymbol{\Theta}(\mathbf{x}_i) & 0
    \end{bmatrix} \begin{bmatrix}
                    {\boldsymbol{\xi}_k} \\
                    {\bf z}
                  \end{bmatrix} = \begin{bmatrix}
                                    \boldsymbol{\Theta}(\mathbf{X}^{(-i)})^T \dot{\mathbf{X}}^{(-i)}_k \\
                                    \dot{\mathbf{x}}_i
                                  \end{bmatrix}.
   \label{eq: KKT equations}
\end{equation}
The solution to this problem yields a surrogate SINDy model $\tilde{\boldsymbol{\Xi}}_i$, which is then compared to the $i$th SINDy model $\boldsymbol{\Xi}_i$ in the ensemble to compute a non-conformity score: 
\begin{equation}
    s_{i}=\|\tilde{\boldsymbol{\Xi}}_i-{\boldsymbol{\Xi}}_i\|_1.
    \label{conformityScoreFeatureCP}
\end{equation}
This non-conformity score is then used then build a $(1 - \alpha)$ prediction interval by computing the appropriate quantile of these scores.

Here, the constrained least squares problem is solved using the thresholded library from the initial unconstrained SINDy problem. On the reduced library, we only perform constrained least squares without further thresholding model coefficients. However, constructing valid confidence intervals after selecting a model from the data is challenging, due to selection bias. Several works have addressed this problem, for example, proposing a framework for valid post-selection inference by reducing the problem to one of simultaneous inference, thereby widening confidence intervals~\cite{berk2013valid}, or by debiasing model selection algorithms to construct valid confidence intervals~\cite{zhang2014confidence,chernozhukov2018double}. More recent work considers a scenario where a family of pretrained models and a hold-out set is given, and proposes a conformal prediction procedure to construct valid prediction intervals for the selected model that minimizes the width of the interval~\cite{liang2024conformal}. Their method avoids the need for additional data splitting by directly ensuring valid coverage after an efficiency-oriented model selection step, which is particularly important in limited data regimes. 
Exploring these ideas and further extending feature-CP with SINDy is an interesting future research direction.

\section{Results} \label{sec:results}

To evaluate the performance of the proposed SINDy with conformal prediction methods, we conduct experiments on a synthetic dataset generated from the classical predator-prey system. Additional results on different datasets, including chaotic dynamical systems, are presented in the appendix. The predator-prey dynamics are governed by the following system of (stochastic) differential equations:
\begin{equation}
    \begin{aligned}
        \frac{d}{dt}x_1(t) &= \alpha x_1(t) - \beta x_1(t) x_2(t) + w_1(t), \\
        \frac{d}{dt}x_2(t) &= \delta x_1(t) x_2(t) - \gamma x_2(t) + w_2(t), \\
        y_1(t) &= x_1(t) + v_1(t), \\
        y_2(t) &= x_2(t) + v_2(t),
    \end{aligned}
    \vspace{10pt}
\end{equation}
where $ x_1(t) $ and $ x_2(t) $ denote the prey and predator population at time $ t $, $w(t)$ the process noise, and $y(t)$ the observations with noise $v(t)$. The coefficient $ \alpha > 0 $ controls the prey's natural growth rate, $ \beta > 0 $ represents the rate at which predators consume prey, $ \gamma > 0 $ is the natural death rate of the predators, and $ \delta > 0 $ governs the rate at which the predator population grows due to predation. We set the coefficients to $[\alpha,\beta,\gamma,\delta]=[1, 0.1, 1, 0.1]$.
The measurement and process noise we add to the system is zero-mean Gaussian in most of our analyses, but we also include one comparison using non-zero-mean gamma-distributed observation noise.  
Since large noise levels are considered, we use the weak form of SINDy to avoid calculating derivatives on noisy data. For simplicity in the context of the main results section in this paper, we define a small library of candidate functions containing only quadratic terms. This small library makes the identification problem more tractable and facilitates visual interpretation and comparison of the results across different conformal prediction methods.

\vspace{10pt}

\subsection{Time series prediction} \label{sec:prediction}

\begin{figure}[t]
    \centering
    \includegraphics[width=1.0\textwidth]{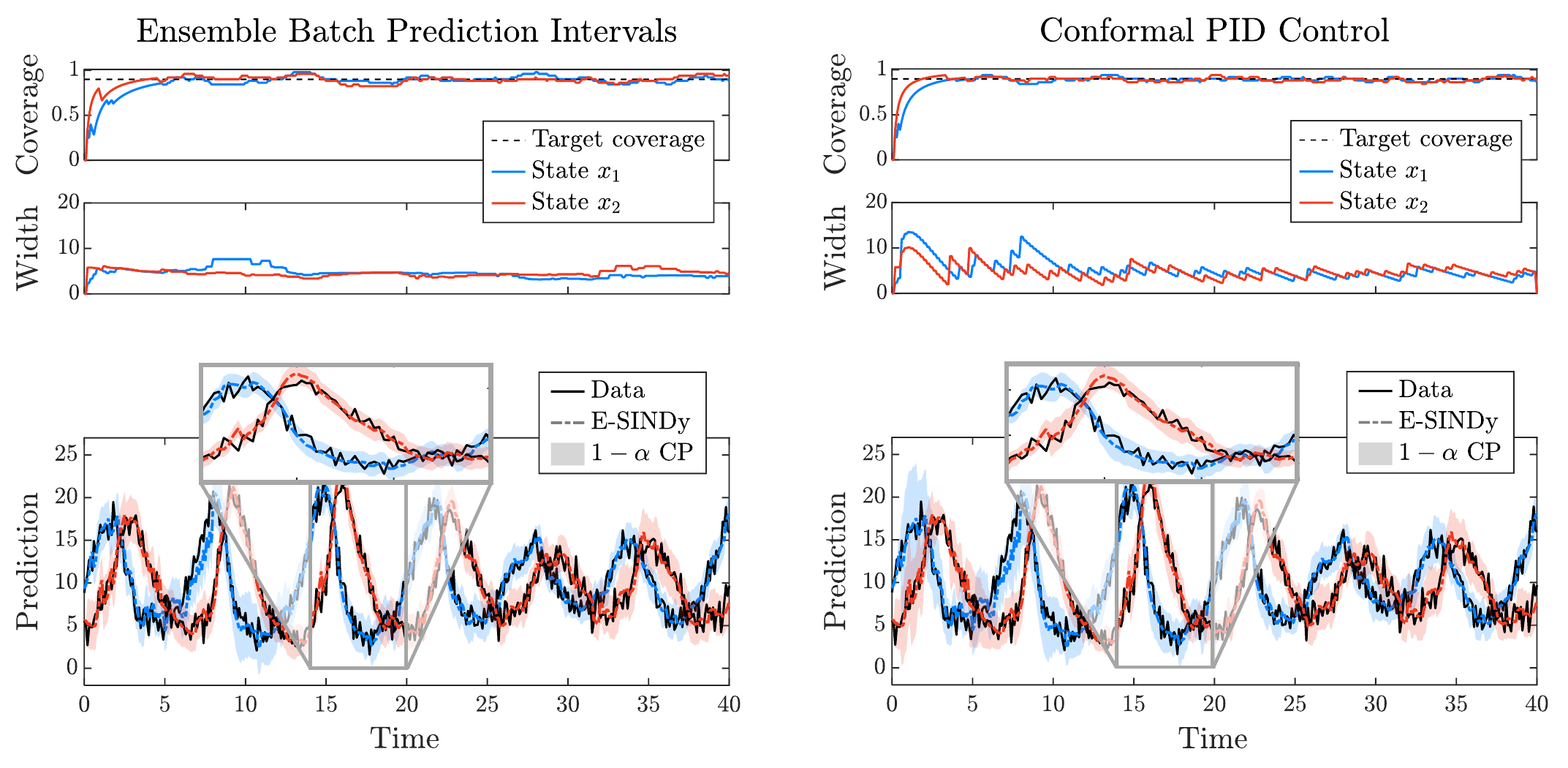}
    \caption[ ] {E-SINDy with conformal prediction for time series prediction using EnbPI and CP-PID. Top: Coverage and width for predator population $x_1$ and prey population $x_2$ over time. Bottom: measurements, predictions, and $90\%$ prediction intervals for predator and prey populations.}
    \label{fig1a}
\end{figure}

We first evaluate the two conformal prediction methods for time series prediction: Ensemble Bootstrap Prediction Intervals (EnbPI) and conformal PID control (CP-PID). The results for both methods are shown in Figure~\ref{fig1a}. 
The top two rows show the coverage (averaged over a trailing window of 50 time steps with $dt=0.1$) and the width of the prediction interval. The bottom row shows the E-SINDy prediction and prediction intervals, along with the observed data. In this example, the prediction horizon (or batch size as called in EnbPI) is two time steps. 
Both approaches successfully achieve the $1-\alpha$ target coverage set to $90\%$ while maintaining relatively narrow prediction intervals, demonstrating effective uncertainty quantification even in a low-data regime.

Notably, CP-PID initially generates wide prediction intervals to reach target coverage due to inaccuracies in the underlying E-SINDy forecasts at the start of the prediction when data is limited. However, it quickly adapts and converges toward the correct coverage level as more data becomes available. In contrast, EnbPI adjusts its prediction intervals more conservatively and exhibits small oscillations around the target coverage, but it avoids rapidly widening the prediction intervals. This difference reflects the more reactive nature of CP-PID compared to the EnbPI calibrated intervals.

\begin{figure}[h]
    \centering
    \includegraphics[width=1.0\textwidth]{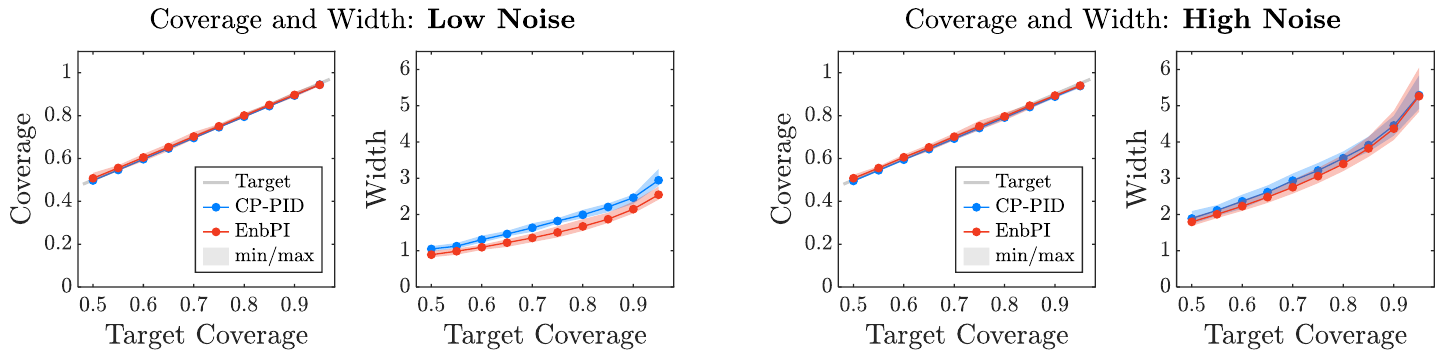}
    \caption[ ] {Coverage and width over target coverage for the predator-prey dynamics using EnbPI and CP-PID, assessed on high and low noise levels.}
    \label{fig1b}
\end{figure}

Figure~\ref{fig1b} shows the coverage and interval widths for both methods across varying target coverage levels, comparing performance under high and low measurement and process noise conditions. In all cases, the methods accurately match the target coverage. Additionally, we observe a nonlinear growth in the width of the prediction intervals as a function of the target coverage level. 
In the appendix, we show additional examples and results for longer prediction horizons and applying both CP time series methods to different chaotic dynamical systems datasets. For all examples, the target coverage is again achieved, although in most cases with wider prediction intervals.

\subsection{Model selection: library feature importance} \label{sec:model discovery}

Next, we evaluate LOCO and LOCO-path using SINDy, the two conformal prediction-inspired approaches for assessing library feature importance. These methods are designed to help select models that best describe the dynamics and to provide uncertainty quantification for the model selection process itself. 
Additionally, the LOCO-path method offers a practical advantage since it avoids the need to select a regularization parameter $\lambda$ required in SINDy.

\begin{figure}[h]
    \centering
    \includegraphics[width=1.0\textwidth]{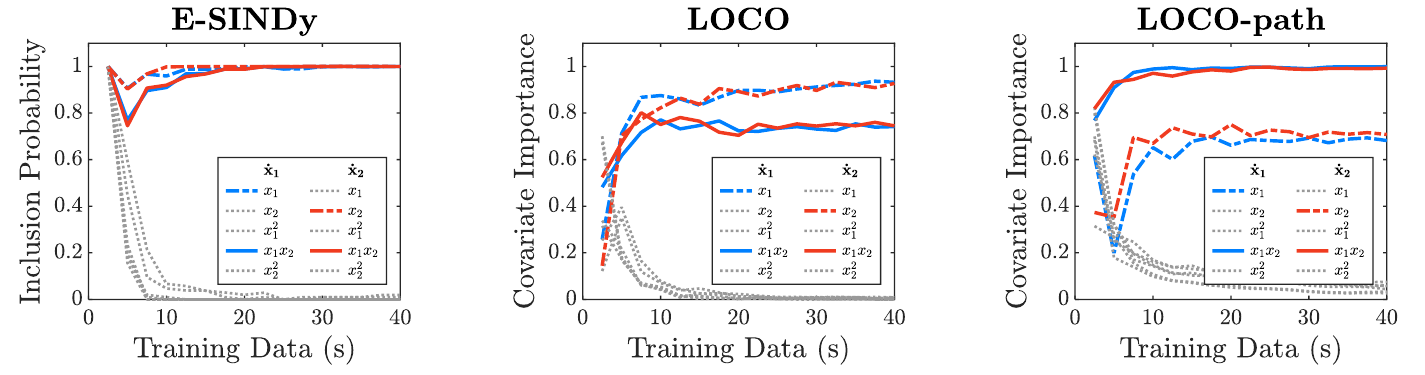}
    \caption[ ] {E-SINDy inclusion probability, LOCO, and LOCO-path results showing a measure of importance for each library feature (covariate), which can inform model selection. Red and blue lines show true active terms, grey lines show non-active terms for the predator-prey dynamics. Inclusion probabilities and importance measures are averaged over 100 noise realizations.}
    \label{fig2}
\end{figure}

The results for both methods are shown in Figure~\ref{fig2}. We also show the baseline E-SINDy inclusion probabilities, which provide a similar measure of variable importance and highlight how often a library feature (or variable) is active in the ensemble of SINDy models. For all three approaches, we show the library feature importance scores (or the inclusion probability) as a function of the training data length used for model selection. We run 100 repeated simulations and plot the averaged scores. Already in the low data regime, the computed importance measures stabilize rapidly. This stabilization allows for clear discrimination between features that contribute to the system dynamics and those that do not. This stabilization is also clearly visible for the inclusion probability results. We see that features corresponding to active terms in the true governing equations consistently exhibit higher importance scores (or inclusion probabilities), while the importance measure of irrelevant library features remains low. These results show that both baseline E-SINDy inclusion probabilities and conformal prediction-inspired methods can be useful in guiding model selection by highlighting relevant features of the underlying system.

\vspace{4pt}

\subsection{Model coefficient uncertainty estimation: feature-CP} \label{sec:param}

\begin{figure}[t]
    \centering
    \includegraphics[width=1.0\textwidth]{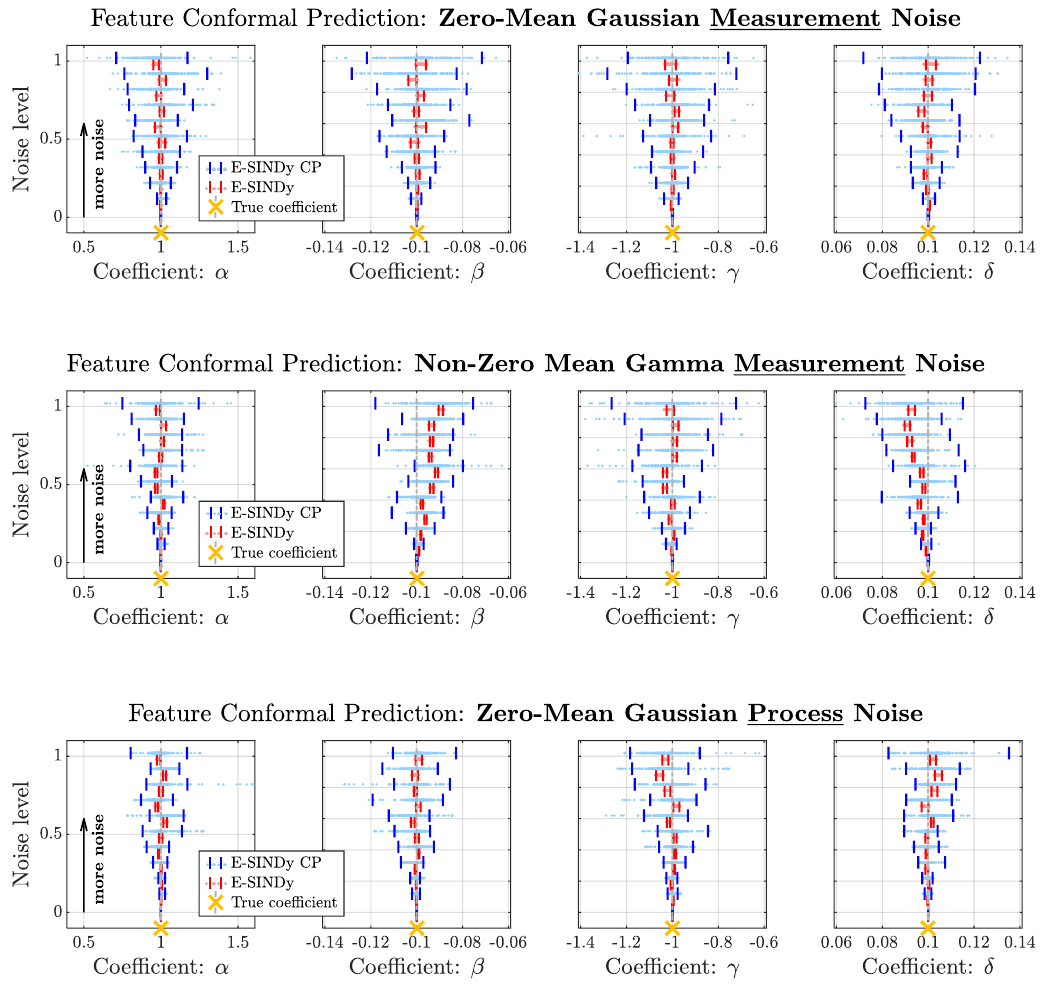}
    \caption[ ] {Feature-CP results evaluating SINDy conformal prediction on model coefficients, for a fixed time series length and varying noise levels. Comparison between standard E-SINDy (red) and E-SINDy with feature-CP (blue), for (top) zero-mean Gaussian measurement noise, (centre) non-zero mean Gamma measurement noise, and (bottom) zero-mean Gaussian process noise. The orange cross and vertical grey dashed lines indicate the true model coefficients used to generate the data (\mbox{$\alpha=1$}, \mbox{$\beta=-0.1$}, \mbox{$\gamma=-1$}, and \mbox{$\delta=0.1$}). The small light blue and light red dots indicate the ensemble model coefficient estimates for standard E-SINDy and E-SINDy CP, and the red and blue vertical bars indicate the corresponding $90\%$ confidence intervals.}
    \label{fig_featureCP}
\end{figure}

In the final set of experiments, we apply the feature-CP method in combination with SINDy to the predator-prey dataset, this time keeping the training data length fixed while systematically increasing the measurement or process noise level. The goal is to assess how conformal prediction can be used to estimate uncertainty in the learned model coefficients under varying noise. The results are shown in Figure~\ref{fig_featureCP}.

We compare the standard E-SINDy coefficient estimates with those obtained using E-SINDy with feature-CP. We visualize each coefficient estimate with small dots, and indicate the corresponding $(1-\alpha)$ confidence intervals, here $90\%$, with vertical bars. Note that we compute the quantiles on the scalar score $s_i$ in Equation~\ref{conformityScoreFeatureCP}, calibrating the intervals on the full model and not on the individual model coefficients. As expected, the uncertainty estimates from E-SINDy are relatively narrow. They generally contain the true coefficient values in the low-noise, zero-mean Gaussian case, but fail to do so reliably when the measurement noise is larger, drawn from a non-zero-mean gamma distribution, or when process noise is added.

In contrast, the feature-CP intervals are more conservative and cover the true coefficients even in the presence of process noise and gamma-distributed measurement noise. These results are encouraging, showing that feature-CP can robustly estimate model uncertainty in noisy settings. However, the resulting intervals are rather wide. Narrower intervals could potentially be estimated by first denoising the measurements before applying feature-CP, following a sequential strategy similar to the denoising step in the SINDy-based approach described in~\cite{kaheman2022automatic}. Other potential extensions could include debiasing~\cite{chernozhukov2018double}, or solving an errors-in-variables problem to consider noise in the time derivatives and the library functions~\cite{bortz2023direct}. In the appendix, we present additional results comparing
feature-CP with Bayesian SINDy (BSINDy)~\cite{fung2025rapid} and an errors-in-variables method (WENDy)~\cite{bortz2023direct}, and we show averaged confidence intervals over multiple noise realizations compared to the single noise realization shown in Figure~\ref{fig_featureCP}.

\section{Discussion and conclusion} \label{sec:discussion}

In this work, we integrated conformal prediction methods with E-SINDy for uncertainty quantification. Results across three applications demonstrate promising performance: achieving desired target coverage for time series forecasting, effectively quantifying feature importance for model selection, and producing uncertainty intervals for identified model coefficients, even under process noise and non-Gaussian measurement noise. Each approach offers different insights into uncertainty quantification and model selection when combined with E-SINDy.

First, we applied conformal prediction methods to time series forecasting with E-SINDy using EnbPI and CP-PID. Both methods achieve the desired target coverage, even in low-data regimes, while maintaining relatively narrow prediction intervals. CP-PID adapts more aggressively, initially generating wide prediction intervals in the low-data limit due to inaccurate E-SINDy predictions, but ultimately converging toward the target coverage with narrow intervals. In contrast, EnbPI adjusts intervals more conservatively, with small fluctuations around the target coverage. A consistent trend in interval width growth is observed with increasing target coverage. 
Second, we explored sparse model selection using conformal prediction-inspired methods. Both the LOCO and the LOCO-path methods with jackknife sampling E-SINDy help quantify feature importance in the SINDy library. Particularly, the LOCO-path method offers the advantage of avoiding the need to tune the regularization parameter required in SINDy. For all methods, the importance measures for true active features stabilize rapidly with increasing data, while irrelevant features are consistently assigned low importance scores. These results indicate that such methods may be useful for dynamical system model selection and feature importance analysis, even in noisy or data-limited regimes. 
However, LOCO and LOCO-path require repeated model retraining, making them computationally expensive, which might constrain their application in real-time or higher-dimensional settings.
Finally, we tested the feature-CP method for quantifying uncertainty in model coefficients under varying levels of measurement and process noise. While standard E-SINDy coefficient intervals are accurate under Gaussian measurement noise, they fail to maintain valid coverage under process noise and non-zero-mean gamma measurement noise. In contrast, feature-CP produces more conservative intervals that successfully cover the true coefficients across noise types. However, this increased robustness comes at the cost of wider intervals. To improve efficiency, future work could explore hybrid approaches, such as sequentially applying a denoising step before conformal prediction, as suggested in recent SINDy literature~\cite{kaheman2022automatic}.

Our results demonstrate the versatility and potential of integrating conformal prediction with SINDy. 
As a next step, these methods should be applied to more challenging real-world datasets to assess their reliability in safety-critical applications. 
In addition, comparing them to traditional ARIMA time series prediction methods~\cite{durbin2012time}, recent SINDy-EKF~\cite{rosafalco2024ekf} data assimilation methods, and Bayesian methods for uncertainty quantification in sparse model discovery~\cite{fung2025rapid} will help evaluate their practical impact. 
While we do not provide formal theoretical results for the SINDy-specific conformal prediction methods, such as finite-sample coverage guarantees or convergence of error bounds, establishing such guarantees for the SINDy library feature importance and feature-CP procedures remains an important direction for future research.

\titleformat{\section}[block]{\large\bfseries}{}{0pt}{}
\renewcommand{\thesection}{\Alph{section}}

\numberwithin{figure}{section}
\numberwithin{equation}{section}

\appendix

\section{\underline{Appendix A} \hspace{6pt} Additional time series prediction results}
\begin{figure}[h]
    \centering
    \includegraphics[width=1.0\textwidth]{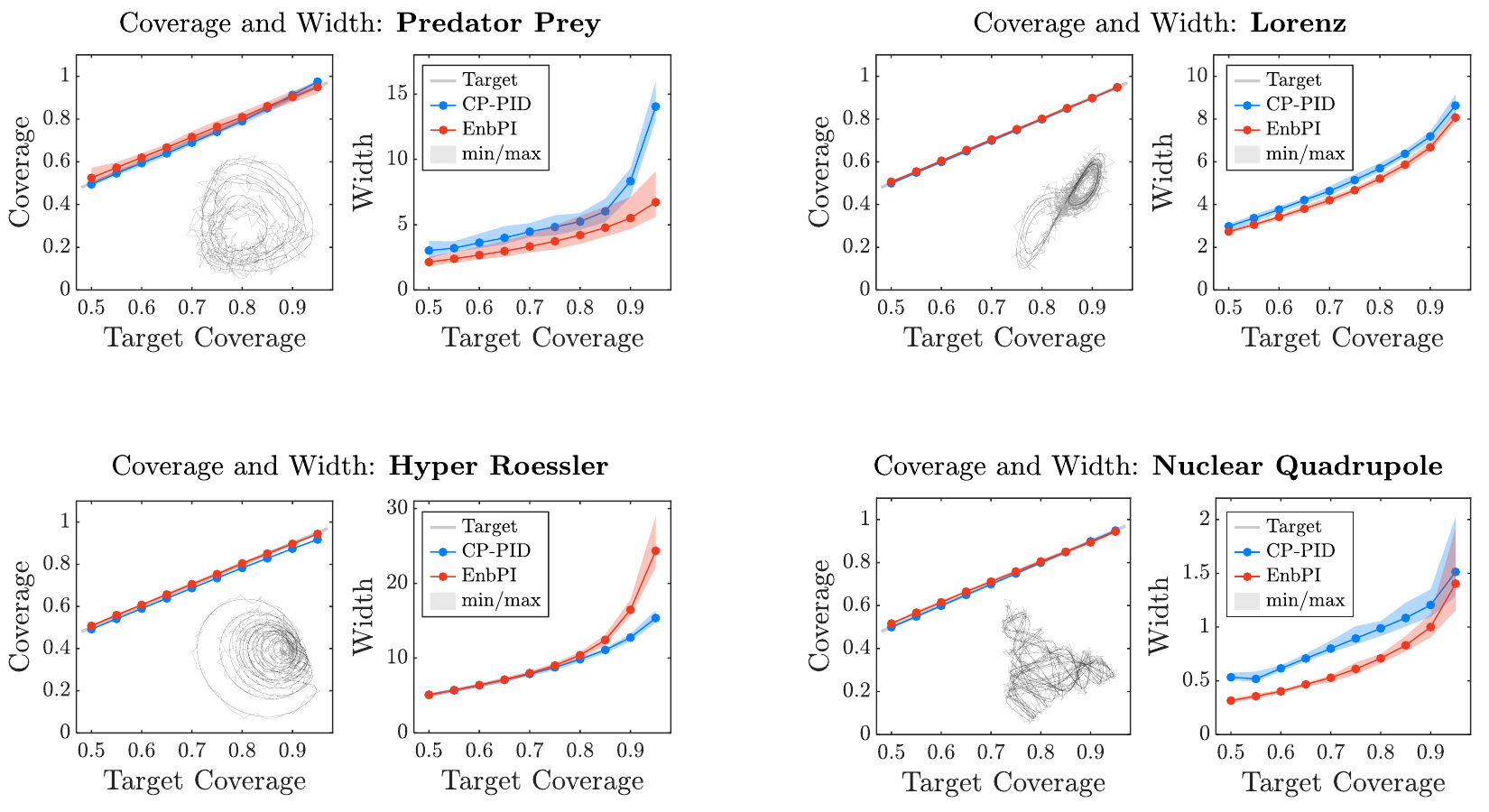}
    \caption{SINDy with EnbPI and CP-PID for longer prediction on predator prey data (top left), and on three chaotic dynamical systems including the Lorenz (top right), Hyper Roessler (bottom left), and Nuclear Quadrupole system (bottom right).}
    \label{fig:appendix1}
\end{figure}

To further evaluate CP methods with SINDy in time series forecasting, we present additional results on the predator-prey system using a longer prediction horizon (10 time steps instead of 2). We also assess performance on three chaotic dynamical systems with varying characteristics~\cite{gilpin2021chaos,kaptanoglu2023benchmarking}: (1) the Lorenz system, a system of three ordinary differential equations with polynomial terms up to second order, commonly used as a benchmark for learning chaotic dynamics; (2) a variant of the Rössler system exhibiting hyper chaos (characterized by two or
more positive Lyapunov exponents); and (3) a quadrupole boson Hamiltonian system with four equations and third-order polynomial nonlinearities. Results are shown in Figure~\ref{fig:appendix1}. 
In all cases, the results indicate that the desired target coverage is achieved, although the prediction intervals are wider, as expected for more complex or longer-horizon predictions. In these experiments, the CP-PID gains were not tuned. We expect that tuning these parameters would enhance performance, potentially through automated methods as proposed in the original CP-PID paper~\cite{angelopoulos2023conformal,bhatnagar2023improved,gibbs2024conformal}.


\section{\underline{Appendix B} \hspace{6pt} Feature-CP compared with WENDy and BSINDy}
\begin{figure}[h]
    \centering
    \includegraphics[width=1.0\textwidth]{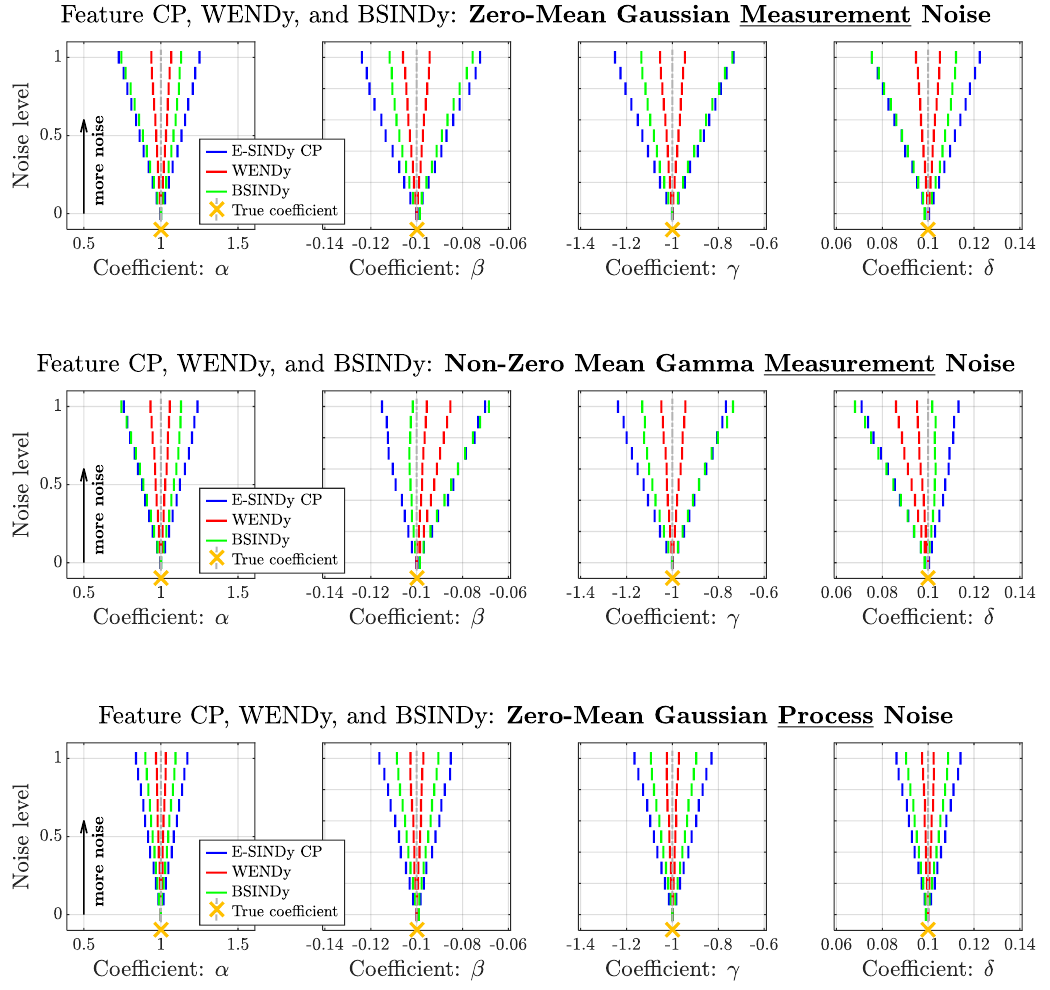}
    \caption{E-SINDy with feature-CP (blue) compared with WENDy (red) and BSINDy (green). Results shown for a fixed time series length and varying noise levels. Each noise level is run on 100 realizations of noise, and the $90\%$ confidence intervals are averaged for plotting. Comparison between the three methods for (top) zero-mean Gaussian measurement noise, (centre) non-zero mean Gamma measurement noise, and (bottom) zero-mean Gaussian process noise. The orange cross and vertical grey dashed lines indicate the true model coefficients used to generate the data, and the vertical bars show the $90\%$ confidence intervals for the three methods.}
    \label{fig:appendix2}
\end{figure}

We briefly compare feature-CP with WENDy and BSINDy. WENDy (Weak-form Estimation of Nonlinear Dynamics) is an errors-in-variables extension of the weak-form SINDy framework that considers noise both on the time derivatives and also the library features~\cite{bortz2023direct}. BSINDy (Bayesian SINDy) is a reformulation of the SINDy method within a Bayesian framework that uses Gaussian approximations for the prior and likelihood. The results are shown in Figure~\ref{fig:appendix2}. 
We use the same predator-prey datasets as in Section 4 for a fixed time series length and varying noise levels, with: 
(1) zero-mean Gaussian measurement noise, (2) non-zero-mean Gamma noise, and (3) zero-mean Gaussian process noise. 
Each noise level is run on 100 realizations of noise, and the $90\%$ confidence intervals are averaged for plotting. The results presented here compare E-SINDy CP with WENDy and BSINDy. WENDy performs well under Gaussian measurement noise. However, under Gamma-distributed measurement noise, we can observe a bias in the model coefficient estimates of the quadratic terms ($\beta$ and $\delta$), and the true model coefficients occasionally fall outside the confidence intervals, especially at higher noise levels. 
We also observe some convergence issues of the WENDy algorithm for the predator-prey data with process noise, although these may be mitigated through careful adaptation of the WENDy implementation. 
The results are encouraging since the confidence intervals produced by WENDy are more reliable and wider than those from standard E-SINDy and narrower than those from E-SINDy CP. BSINDy also performs well under all three noise conditions, although the intervals are rather wide, similar to the ones predicted by E-SINDy CP, and we also observe a bias for the quadratic terms at higher noise levels. Exploring the integration of both BSINDy and the WENDy errors-in-variables formulation with E-SINDy CP remains a promising direction for future research.

\bibliographystyle{unsrt}
 \begin{spacing}{.91}
 \setlength{\bibsep}{2.pt}
 \bibliography{references}
 \end{spacing}

\end{document}